\documentclass{article} 
\usepackage[
backend=biber,
style=numeric,
]{biblatex}
\title{A bibLaTeX example}
\usepackage{graphicx}%
\usepackage{multirow}%
\usepackage{amsmath,amssymb,amsfonts}%
\usepackage{amsthm}%
\usepackage{mathrsfs}%
\usepackage[title]{appendix}%
\usepackage{xcolor}%
\usepackage{textcomp}%
\usepackage{manyfoot}%
\usepackage{booktabs}%
\usepackage{algorithm}%
\usepackage{algorithmicx}%
\usepackage{algpseudocode}%
\usepackage{listings}%
\usepackage{array}
\usepackage{authblk}
\usepackage{color}
\usepackage{tabularx}
\usepackage{placeins}

\title{CLOCR-C: Context Leveraging OCR Correction with Pre-trained Language Models}
\author{Jonathan Bourne}
\affil{Centre for Advanced Spatial Analysis, University College London}

\begin{document}

\maketitle

\begin{abstract}
The digitisation of historical print media archives is crucial for increasing accessibility to contemporary records. However, the process of Optical Character Recognition (OCR) used to convert physical records to digital text is prone to errors, particularly in the case of newspapers and periodicals due to their complex layouts. This paper introduces Context Leveraging OCR Correction (CLOCR-C), which utilises the infilling and context-adaptive abilities of transformer-based language models (LMs) to improve OCR quality. The study aims to determine if LMs can perform post-OCR correction, improve downstream NLP tasks, and the value of providing the socio-cultural context as part of the correction process. Experiments were conducted using seven LMs on three datasets: the 19th Century Serials Edition (NCSE) and two datasets from the Overproof collection. The results demonstrate that some LMs can significantly reduce error rates, with the top-performing model achieving over a 60\% reduction in character error rate on the NCSE dataset. The OCR improvements extend to downstream tasks, such as Named Entity Recognition, with increased Cosine Named Entity Similarity. Furthermore, the study shows that providing socio-cultural context in the prompts improves performance, while misleading prompts lower performance. In addition to the findings, this study releases a dataset of 91 transcribed articles from the NCSE, containing a total of 40 thousand words, to support further research in this area. The findings suggest that CLOCR-C is a promising approach for enhancing the quality of existing digital archives by leveraging the socio-cultural information embedded in the LMs and the text requiring correction.
\end{abstract}

Print media archives are increasingly digitising their records to increase accessibility \cite{terras_rise_2011}. Access to digital historical archives is particularly important when it comes to the accessibility of contemporary historical records, such as periodicals and news media, which give insight into views and opinions of historical events when they occurred. However, the process of Optical Character Recognition (OCR), which is used to convert physical records to digital type, is prone to error \cite{smith_research_2018}, particularly in the case of newspapers and periodicals which have more complex layouts \cite{chiron_impact_2017}. Such errors can negatively impact the quality of research using those archives \cite{traub_impact_2015, chiron_impact_2017}. Given the amount of OCR documents already in existence post-OCR correction and assessing OCR quality are areas of active research \cite{chiron_icdar2017_2017, van_strien_assessing_2020, nguyen_survey_2021, neudecker_survey_2021}. 

There are several popular approaches to post-OCR correction; one of the more successful approaches has been crowd-sourcing the correction with online platforms \cite{holley_many_2009, suissa_toward_2020, mechaca_web_2021} or using online security tests \cite{von_ahn_recaptcha_2008}. However, due to its success in other fields and increased computational power, machine learning approaches are increasingly being used \cite{ soper_bart_2021, karthikeyan_ocr_2022, amrhein_supervised_2018, mokhtar_ocr_2018, mei_statistical_2018, ramirez-orta_post-ocr_2022}. These approaches hope to take advantage of the opportunities provided by large corpora to create statistical algorithms able to cheaply increase the speed and quality of post-OCR correction. However, such attempts are not without difficulty; the 2017 ICADAR post-OCR correction competition found that only about half the submitted methods were able to improve the OCR quality \cite{chiron_icdar2017_2017}. 

A promising area of development for post-OCR processing is the transformer architecture for Language Models (LMs)  \cite{vaswani_attention_2017}.
The introduction of the transformer architecture in 2017 sparked rapid development in Natural Language Processing, with increasingly capable LMs being produced able to perform tasks at human or even superhuman levels \cite{wang_superglue_2019, zhang_retrospective_2020, he_deberta_2021, xu_human_2022}. These rapid gains were related to the transformer’s ability to ‘pay attention’ to different parts of the text. By tying together long-range dependencies in a way that had not been previously possible, it significantly enhanced the comprehension of text. 
Another important development was Masked Language Modelling (MLM), created as a response to the challenges of training the bi-directional BERT model \cite{devlin_bert_2019}.
MLM, inspired by a test of text readability \cite{taylor_cloze_1953},  follows a process in which some fraction of the tokens are randomly masked, and the algorithm has to `Infill' the gap with the correct token. This approach to training LMs produced much more valuable representations, making the off-the-shelf BERT model a generalist that could be fine-tuned on a small amount of data to outperform many specific models, which the authors demonstrated by reaching state-of-the-art on 11 different benchmarks. 

Since the release of BERT in 2019, the most high-performing models use the ``autoregressive'' architecture (focusing on next token prediction) instead of the bi-directional architecture used by BERT. In addition, the size and power of the LMs have grown dramatically, with the focus now on models with Billions of parameters compared to BERT's 110M. The number of parameters is usually proportional to the capability of the LM, with more parameters getting higher scores on the various LM benchmarks \cite{kaplan_scaling_2020} that are currently being used (examples \cite{zellers_hellaswag_2019, sakaguchi_winogrande_2019,hartvigsen_toxigen_2022, hendrycks_measuring_2021}). Although there is debate as to whether the total parameters, the number of training tokens \cite{hoffmann_training_2022}, or even the quality of dataset \cite{gunasekar_textbooks_2023}, is the most important aspect of LM performance. In 2022, OpenAI released ChatGPT web-based interface to a pre-trained multi-lingual language model GPT-3 \cite{brown_language_2020}; they also provided API access to allow developers programmatic access to the model. The `magic' of GPT was that as a generative model, users could type in questions or commands, also known as `prompts', which would return informative human-like responses. The release of ChatGPT sparked a flurry of pre-trained LMs available online and through API with capabilities that far exceeded the training budgets available to the majority of users. These models, trained on a vast range of social, cultural, and historical topics, can tie together information from across a piece of written text, a capability shared by all transformer models.

\subsection{Introduction to CLOCR-C}
\label{sect_intro_to_clocr}

It is the combination of powerful pre-trained LMs and their ability at `infilling' that is of interest in this paper. As seen in Table \ref{tab:knowledge_types}, the corrupted text of poor quality OCR is conceptually similar to the masking process as the LM must correctly predict the missing word or words. The infilling capacity of LMs is already being used in humanities research in the form of the recovery of ancient texts \cite{lazar_filling_2021, fono_embible_2024}. The goal of the task presented to the model is to identify the corrupt text and infill the corrupt `masks' to correct the text to its original state. This process is effectively editing the provided text to create a human-readable text that maximises the likelihood of the resultant text distribution. The prior distribution of the pre-trained LM is shaped by the prompt, the additional context provided by the words in the OCR text to be corrected, and the inferred socio-cultural context. In this sense, the LMs leverage the contextual information related to the text to inform the post-OCR correction. As such, this paper defines the term Context Leveraging OCR Correction (CLOCR-C).

Conceptually, CLOCR-C can be thought of as looking for clues in OCR text that may be used to help correct elements that are missing or erroneous due to the OCR process. Reconstructing the most likely original text requires high levels of reading comprehension, as well as commonsense, cultural, historical, and social knowledge. Table \ref{tab:knowledge_types} provides examples of different sentences and the types of possible clues available and the knowledge required to complete the text. With this range of challenges the task of CLOCR-C appears similar to the challenge presented by the multi-part reading benchmark SuperGLUE \cite{wang_superglue_2019} particularly the test elements taken from MultiRC \cite{khashabi_looking_2018}, ReCord \cite{zhang_record_2018} and Winograd \cite{levesque_winograd_2012}. \footnote{Since its release in 2019, SuperGLUE has been broadly surpassed. As of this writing, the top model on the leaderboard is a 6B parameter model \cite{zhong_toward_2022} that has held the spot since 2022. More recent models no longer use this benchmark.}

\begin{table*}[ht]
    \caption{Types of knowledge and reasoning for different question-answer pairs. Clues are shown in blue, the missing text is shown as red stars `*'}
    \centering
    \begin{tabular}{p{0.4\linewidth}c p{0.2\linewidth}}
        \toprule
        \textbf{Phrase} & \textbf{Answer} & \textbf{Knowledge} \\
        \midrule
        The \textcolor{red}{\textbf{***}} boarded the boat. It was the largest \textcolor{blue}{\textbf{seal}} I ever saw. & seal & Co-reference resolution \\
        \hline
        It was h\textcolor{red}{\textbf{**}} first time being saluted as a captain. \textcolor{blue}{\textbf{Mary}} had only just been promoted. & her & Gender/pronoun agreement \\
        \hline
        The most common way to get goods to market is by \textcolor{red}{\textbf{***}}. & truck & Common sense reasoning \\
        \hline
        When \textcolor{blue}{\textbf{Oscar Wilde}} was young the most common way to get goods to market was by \textcolor{red}{\textbf{***}}. & cart & Historical and Cultural\\
        \bottomrule
    \end{tabular}
    \label{tab:knowledge_types}
\end{table*}

\subsection{Objectives and contributions of the paper}
This paper introduces the concept of Context Leveraging OCR Correction (CLOCR-C), which uses transformer-based large language models' infilling and context adaptive abilities to improve OCR quality. It seeks to answer three questions. 

\begin{itemize}
    \item Can LMs improve the accuracy of OCR outputs in newspapers and periodicals? (Primary research question)
    \item Does the post CLOCR-C error rate improve downstream NLP tasks?
    \item Does the inclusion of the socio-cultural context improve the accuracy of OCR outputs?
\end{itemize}

This paper is not the first to have attempted to use LMs in this way, \textcite{boros_post-correction_2024} tested 14 LMs on eight different datasets and found that in no case could the LMs improve the quality of the OCR. Although not an auspicious start, this paper aims to prove otherwise.

\section{Methodology}

The methodology is broken into three sections, in which the data set and sampling method, language models and experiments and evaluation methods are each defined and described, in turn.

\subsection{Data}
\label{sect:data}

This paper uses three datasets of digitally archived newspapers from the UK, Australia, and the USA. These datasets are, a newly released dataset the 19th Century Serials Edition \cite{brake_nineteenth-century_2008}, and two datasets which are part of the Overproof collection \cite{evershed_correcting_2014}, The Sydney Morning Herald and Chronicling America.

\subsubsection{The Nineteenth-Century Serials Edition}

The Nineteenth-Century Serials Edition (NCSE) \cite{brake_nineteenth-century_2008} is a newly available open-source archive of 6 periodicals, which have been digitally scanned and had optical character recognition performed on them. The periodicals in the archive are The Monthly Repository and Unitarian Chronicle, Northern Star, The Leader, The English Woman's Journal, The Tomahawk, and the Publisher's Circular. Table \ref{tab:periodicals_table} shows each newspaper's date range for which data is available, the number of issues, and the number of articles. 

\begin{table*}
\caption{The periodicals in the NCSE and their key information}
\label{tab:periodicals_table}
\begin{tabular}{p{4cm}lrrr}
\toprule
Title & Years & Issues & Articles (k) & Tokens (M) \\
\midrule
Monthly Repository and Unitarian Chronicle & 1806-1837 & 487 & 51 & 25 \\
Northern Star & 1837-1852 & 2201 & 231 & 252 \\
Leader & 1850-1860 & 1011 & 162 & 88 \\
English Woman’s Journal & 1858-1864 & 91 & 8 & 4 \\
Tomahawk & 1867-1870 & 188 & 13 & 3 \\
Publishers’ Circular & 1880-1890 & 285 & 57 & 42 \\
\bottomrule
\end{tabular}
\end{table*}

An example of the raw OCR from the dataset and its transcription of the same article is shown in Table \ref{tab:example_text}. The table shows several different error types that appear in the raw OCR and highlights two particularly challenging examples where the OCR mixes the physical lines of print, resulting in part words in different parts of the text. Despite this, as discussed in Section \ref{sect_intro_to_clocr}, the interested reader may, through careful examination and some knowledge of 19th-century Britain, be able to reconstruct the text to a passable degree.

\begin{table*}[h]
    \caption{Example of the raw OCR and the transcribed text from an advert from an 1868 edition of the Tomahawk. Errors cause by the mixing of printed lines during the original OCR process are highlighted in red and blue.}
    \centering
    \begin{tabular}{p{0.45\textwidth}p{0.45\textwidth}}
        \toprule
         \textbf{Raw OCR} & \textbf{Transcribed text} \\
        \midrule
        \texttt{C * RYST AL PALACE . —The }\textcolor{red}{\texttt{VATE}} \texttt{\^{}— *¦ Magnificent DINING-ROOMS SUXTE which of PUBLIC have been and re-deco }\textcolor{red}{\texttt{PRI}}\texttt{- - rared by Messrs , Jackson , and Graham , is NOW OPEW . mic Bre Deieuners akfasts art . served / Banqiiets in the , Private highest Din style >> ers of , an the < l gastrono Wcddirjj - tages "Whitebait . in perfection . Wines of the choicest vin}\textcolor{blue}{\texttt{partment}}\texttt{ BERTRAM . and ROBERTS , Refreshment \_ }\textcolor{blue}{\texttt{De}}\texttt{-} &
        \texttt{CRYSTAL PALACE.-The Magnificent SUITE of PUBLIC and }\textcolor{red}{\texttt{PRIVATE}}\texttt{ DINING-ROOMS, which have been re-decorated by Messrs. Jackson and Graham, Is NOW OPEN . \newline \newline Dejeuners, Banquets, Private Dinners, and Wedding Breakfasts served in the highest style of the gastronomic art. \newline \newline Whitebait in perfection. Wines of the choicest vintages. BERTRAM and ROBERTS, Refreshment }\textcolor{blue}{\texttt{Department}} \texttt{.} \\
        \bottomrule
    \end{tabular}
    \label{tab:example_text}
\end{table*}

\subsubsection{The Overproof collection}

\textcite{evershed_correcting_2014} introduced the Overproof collection of three datasets for testing post-OCR correction of newspaper articles. Dataset 1 is taken from the TROVE \cite{holley_many_2009} collection and the OCR transcriptions created using crowd-sourcing. However, \textcite{evershed_correcting_2014} state that this dataset has significant quality issues resulting from the crowd-sourced approach. These issues included not all lines being corrected, line boundaries being changed, content being added, and words being changed. As a result, they found that the dataset could not be reliably used to measure the performance of a post-OCR correction system. To resolve this issue, they created Dataset 2 and Dataset 3, which have high-quality line-aligned transcriptions. Dataset 2 is a subset of 159 articles from dataset 1, and dataset 3 is a collection from an American newspaper archive \cite{humanities_chronicling_2005}. Because of the issues raised by \textcite{evershed_correcting_2014}, and in contrast to \cite{van_strien_assessing_2020, boros_post-correction_2024}, this paper will use datasets two and three. Details of the two datasets are shown in table \ref{tab:overproof_dataset_stats}. Using the Overproof collection has a second advantage as it also includes the results of the ``Overproof" post-OCR correction algorithm, after which it is named and which acts as a secondary benchmark baseline.

\begin{table*}
\caption{Summary Statistics of Overproof by dataset}
\label{tab:overproof_dataset_stats}
\begin{tabular}{p{5cm}cccc}
\toprule
Title & Articles & Words & Tokens & Mean Tokens \\
\midrule
Sydney Morning Herald & 159 & 52640 & 80992 & 509 \\
Chronicalling America & 46 & 18292 & 25178 & 547 \\
\textbf{Total} & \textbf{205} & \textbf{70932} & \textbf{106170} & \textbf{518} \\
\bottomrule
\end{tabular}
\end{table*}

\subsubsection{Sampling methodology and quality metrics}
\label{sect:sampling}

The Overproof data will be used in it's entirety, however the NCSE dataset is too large and needs to be sampled. In order to have a representative gold label test-set that can accurately reflect the distribution of the data in the NCSE, care needs to be taken. The periodicals cover almost 100 years of technological innovation and societal change; as such, it cannot be assumed that the style of language, layout or production quality of the newspapers remains constant.

Due to the challenges associated with article-level transcription, it was more practical to transcribe entire pages of text than randomly sampled articles. As such, the dataset was created by performing stratified random sampling at the page level. Manual inspection of test corrections showed that certain simple heuristics could be used as an indicator of scan quality, with some scans producing unrecoverable junk due to either high levels of noise or structural errors in the scanning process (e.g. line mixing), quality could be improved by setting the maximum ratio of symbols to tokens ratio where symbols are the character set (\texttt{: , ; , - , \_ , + , * , \textasciicircum , \textbar , \textbrokenbar , ' , ! , / , > , ] , [, >>}). This symbol to token ration seems to work, because poor quality OCR introduces a much higher incidence of symbols than normal text (see Table \ref{tab:example_text}. The difference in the distribution of symbols between the periodicals was significant and related to the number of adverts and lists, as such articles in the top 10\% of symbol-to-token ratios by periodical were dropped. In addition, for transcription efficiency only pages with a minimum of 500 tokens were included, this also prevents pages which are primarily images being used.

Statistics on the transcribed dataset are shown in Table \ref{tab:final_dataset_stats}.

\begin{table*}[h]
\caption{Summary Statistics of the dataset by periodical}
\label{tab:final_dataset_stats}
\begin{tabular}{p{5cm}cccc}
\toprule
Title & Articles & Words & Tokens & Mean Tokens \\
\midrule
Monthly Repository and Unitarian Chronicle & 17 & 5594 & 6952 & 409 \\
Northern Star & 15 & 13237 & 16329 & 1089 \\
Leader & 9 & 7946 & 9983 & 1109 \\
English Woman’s Journal & 8 & 4035 & 5038 & 630 \\
Tomahawk & 30 & 5775 & 8698 & 290 \\
Publishers’ Circular & 12 & 4125 & 7376 & 615 \\
\textbf{Total} & \textbf{91} & \textbf{40712} & \textbf{54376} & \textbf{598} \\
\bottomrule
\end{tabular}
\end{table*}

Once transcribed, the pages were matched with the segmented articles; this step is crucial for the process as the transcribed text was to be matched with the raw OCR, not the other way around. As a result, if the OCR of an article has no title or misses the first sentence, this information from the transcription is not included. This choice has been made as it is not possible to post-OCR correct text, which is not at all present, and this paper is not exploring broader questions of OCR quality; this avoids some of the issues faced by the Overproof dataset \cite{evershed_overproof_2014}. Only articles that contain at least 100 tokens were included; this is to avoid using article titles or extremely short adverts. Each article is saved as an individual file with a unique identifier.

This dataset distinguishes itself from many of the other post-OCR correction datasets \cite{evershed_correcting_2014, chiron_icdar2017_2017, rigaud_icdar_2019, strobel_improving_2019, romanello_optical_2021, platanou_handwritten_2022}, because it has no `text alignment'. Instead, the entire OCR article is considered a single semantically coherent piece. This paper contends that providing the entire article allows better leveraging of contextual information and prevents issues arising due to text alignment that do not affect correction quality. However, it requires that the models have either a sufficiently large context window or that the text is processed in chunks.

\subsection{Language Models used in the study}

This paper compares eight popular LMs for post-OCR correction; the models are GPT-4 \cite{openai_gpt-4_2023}, GPT-3.5 \cite{brown_language_2020}, Llama 3 \cite{aimeta_llama_2024}, Gemma \cite{mesnard_gemma_2024}, Mixtral 8x7b \cite{jiang_mixtral_2024}, Claude 3 Opus \cite{anthropic_claude_2024}, and Claude 3 Haiku \cite{anthropic_claude_2024}, details on these models can be seen in Table \ref{tab:LM_types}. The models were chosen to cover the largest LM companies, and which were available through a hosted API. At the time of writing, Google's top model, Gemini, was unavailable and could not be included. As can be seen from Table \ref{tab:LM_types}, model size/parameter count varies greatly. As can be seen, the context windows shown in Table \ref{tab:LM_types} of the models shown are far more than the mean article sizes shown in Table \ref{tab:final_dataset_stats}.

\begin{table*}
    \caption{Details on the LMs explored in this paper. Models with * after the name are open-source}
    \centering
    \begin{tabular}{lccc}
        \toprule
        \textbf{Model}  & \textbf{Parameters} &  \textbf{Context (k)} &  \textbf{model id} \\  
        \midrule
        Claude 3 Haiku & Unknown & 200   & claude-3-haiku-20240307 \\
        Claude 3 Opus & Unknown & 200    &  claude-3-opus-20240229\\
        Gemma* & 7B &  8     & gemma-7b-it \\ 
        GPT-3.5 & Unknown\footnotemark[2]  &  16   & gpt-3.5-turbo\\ 
        GPT-4 & Unknown\footnotemark[1] &  128  & gpt-4-turbo-preview \\
        Llama 3*  &  70B &  4  & meta-llama-3-70b-instruct\\ 
        Mixtral 8x7B* & 56B &  32  & mixtral-8x7b-32768 \\ \bottomrule
    \end{tabular}
    \label{tab:LM_types}
\end{table*}
\footnotetext[1]{Rumoured to be 1.75 Trillion \cite{schreiner_gpt-4_2023}}
\footnotetext[2]{Assumed to be at least 175 Billion \cite{hendrycks_measuring_2021}}

\subsection{Experimental setup}

This subsection is again subdivided into three sections. The first section describes the prompts used in the prompt selection process and the LM evaluation metrics. Then, the experiments for testing improvement on downstream tasks are explained. Finally, a simple conceptual demonstration of the Context leveraging used by LMs to perform post-OCR correction and follow-up experiment is provided. 

\subsubsection{Prompt variations and selection process}

A key aspect of using LMs is writing a prompt that produces results most aligned with your intention. This aspect of using language models is known as ``prompt engineering", and how to optimise a prompt is emerging as its own field \cite{white_prompt_2023, sorensen_information-theoretic_2022, wang_prompt_2024}. The objective of this study is to establish a performance baseline, not to optimise the prompts themselves, although some degree of prompt engineering is of course necessary.

This paper uses a modular prompt format of increasing length and complexity; the available sub-prompts are shown in table \ref{tab:instructions}. These sub-prompts are converted to prompts as shown in table \ref{tab:combined_prompts}. The prompts are tested using the dev-set described in section \ref{sect:sampling}. 
It has been suggested that for long texts, it can be more effective to put the prompt after the text instead of the more conventional before. To minimise calculation this paper will assume that having the prompt in the system message and before the OCR text produces a similar result. As such, two prompt formats will be used: prompt as a system message and prompt after the OCR text. When prompt and text are supplied together, the form will be LMs is \verb|"{OCR text}" + "\n\n" + {prompt}|, that is, the OCR text followed by a double line break followed by the prompt. For this test, the four closed-source models will be used; this is to get models from two different companies and of at least two significantly different sizes. As such, the total number of groups tested will be eight prompts, two prompt formats, and four models, making 64 sets of tests. As there may be a range of outcomes across the models, the top 2 performing prompts will be used to evaluate the LMs on the three test sets described in section \ref{sect:data}. The dev-set used in this process is a small collection of 20 articles containing 5000 words.

\begin{table*}[ht]
\centering
\caption{Sub-prompts for Post-OCR Text Recovery}
\label{tab:instructions}
\begin{tabularx}{\textwidth}{l|X|l}
\toprule
\textbf{Sub-prompt name} & \textbf{Description} & \textbf{ID}\\ \midrule
Basic& Please recover the text from the corrupted OCR. & a\\ \hline
Expertise  & You are an expert in post-OCR correction of documents. & b \\ \hline
Recover& Using the context available from the text please recover the most likely original text from the corrupted OCR. & c\\ \hline
Publication Context& The text is from an English newspaper in the 1800's. & d \\ \hline
Text Context & The text may be an advert or article and may be missing the beginning or end. & e \\ \hline
Additional Instructions & Do not add any text, commentary, or lead in sentences beyond the recovered text. Do not add a title, or any introductions. & f\\ \bottomrule
\end{tabularx}
\end{table*}

\begin{table}[ht]
\centering
\caption{Combined Prompts for Post-OCR Correction Tasks}
\label{tab:combined_prompts}
\begin{tabular}{l|l}
\toprule
\textbf{Combined Prompt Label} & \textbf{Sub-prompts} \\ \midrule
basic prompt & a \\ 
expert basic & b + a \\ 
expert recover & b + c \\ 
expert recover publication & b + c + d \\ 
expert recover text prompt & b + c + e \\ 
expert recover publication text & b + c + d + e \\ 
expert recover instructions & b + c + f \\ 
full context & b + c + d + e + f \\ \bottomrule
\end{tabular}
\end{table}

This paper evaluates the performance of the LMs using Character Error Rate (CER), which is calculated using
\begin{equation}
    \textrm{CER} = \frac{S + D + I}{S + D + C}
\end{equation}
where S is the number of substitutions, D is the number of deletions, I is the number of insertions, and C is the correct total number, the denominator is equivalent to the total number of characters in the ground truth reference document. As this paper is interested in whether LMs can improve the quality of the raw OCR baseline, they will all be shown in terms of Error Reduction Percentage (ERP). The ERP is defined as 

\begin{equation}
 \textrm{ERP} = \frac{\textrm{CER} _{\textrm{orig}}-\textrm{CER}_{\textrm{LM}}}{\textrm{CER} _{\textrm{orig}}}\cdot 100   
\end{equation}
where $\textrm{CER}_{\textrm{orig}}$ is the error of the original OCR and $\textrm{CER}_{\textrm{LM}}$ is the error after being corrected by the LMs. The quality of the OCR varies significantly across the corpus, meaning the post-OCR is likely to have a skewed distribution; as such, the median will be used, preventing a few documents from skewing the overall performance. The ERP will be calculated as the micro median, that is that the median ERP will be shown not the ERP of the median CER.

\subsubsection{Downstream task evaluation: Named Entity Recognition (NER)}
\label{sect:meth_ner}

Previous research has shown that poor OCR degrades the quality of downstream tasks such as Named Entity Recognition (NER) \cite{van_strien_assessing_2020, chiron_impact_2017}. While improved CER generally indicates better text quality, it doesn't guarantee accurate recovery of the named entities. Table \ref{tab:NER_errors} shows that CER can be significantly improved, however named entities may still be incorrectly recovered. As such when performing CLOCR-C NER improvement should be measured directly.

\begin{table*}
    \centering
    \begin{tabular}{ccccc}
    \toprule
        Original & Corrupted OCR & Corrected & Original CER & Corrected CER\\
        \midrule
        Jane Austen & Jar.e Aost n & Jane Austin & 0.5 & 0.09 \\
        Duke of Wellington & ..Du k3 0f W3ll1nglgss & Duke of Wellylegs & 0.5 & 0.35\\
        Ada Lovelace & AcIa L.oVe>lace &  Ada Loveslace & 0.33 & 0.08\\
        \bottomrule
    \end{tabular}
    \caption{Examples demonstrating that improvements to CER do not guarantee that Named Entities are correctly recovered.}
    \label{tab:NER_errors}
\end{table*}

To perform NER, this paper uses \verb|Gladiator/microsoft-deberta-v3-large_ner_conll2003| \cite{ingle_gladiatormicrosoft-deberta-v3-large_ner_conll2003_2022}, a fine-tuned Deberta model \cite{he_debertav3_2023}, which is currently the top-ranked model on papers with code for the CoNLL 2003 NER dataset \cite{tjong_kim_sang_introduction_2003}.

Typically, NER is evaluated using the F1 score. However, F1 is sensitive to the position of entities in the text. When performing post-OCR correction, the text length may change due to word substitutions or omissions, which may not materially affect the semantic meaning of the text or the presence and distribution of the named entities. Table \ref{tab:example_sentences} shows how two sentences can have the same meaning and have the same entities but the sentences can look very different, in this case the F1 score matching entity position between the two sentences would be 0 erroneously indicating that the sentences do not contain the same entities. To address this limitation, this paper will use Cosine Named Entity Similarity (CoNES). 

\begin{table*}
\centering
\caption{Example sentences demonstrating entity position changes without semantic difference. The entities are shown in bold }
\label{tab:example_sentences}
\begin{tabular}{p{0.45\linewidth}|p{0.45\linewidth}}
\hline
\textbf{Sentence 1} & \textbf{Sentence 2} \\
\hline
The train, dog, and \textbf{Jim} left \textbf{London}. & The locomotive, the old dog, and of course, \textbf{Jim}, in a cloud of smoke, left \textbf{London}. \\
\hline
\textbf{Entities:} Jim (21st to 23rd characters), London (30th to 35th character) & \textbf{Entities:} Jim (55th to 57th characters), London (86th to 91st characters) \\
\hline
\end{tabular}
\end{table*}

CoNES is a specific application of cosine similarity for measuring the similarity of two texts in terms of the entities irrespective of the position within the text. CoNES is calculated by constructing a vector of the unique named entities in each text, where the value of these elements is the occurrence count of the entities. The CoNES score is the cosine similarity of those two vectors, with 1 meaning the corrected text has all the same entities in matching quantities and 0 being no matching entities. 

To calculate CoNES, first, a vector is constructed for each text, where each element represents a unique named entity identified in both texts. The value of each element is the count of that specific entity in the respective text. The set of unique named entities found in both texts is defined as $E = \{e_1, e_2, ..., e_n\}$. 
$v_p = (p_1, p_2, ..., p_n)$ is the vector representing the predicted entities, where $p_i$ is the count of entity $e_i$ in the predicted text. 
$v_r = (r_1, r_2, ..., r_n)$ is the vector representing the reference entities, where $r_i$ is the count of entity $e_i$ in the reference text.

CoNES is then calculated as the cosine of the angle between these two vectors as shown in Equation \ref{eq:CoNES}

\begin{equation}
\textrm{CoNES} = \frac{v_p \cdot v_r}{||v_p|| \times ||v_r||}
\label{eq:CoNES}
\end{equation}

Applied to the example in Table \ref{tab:example_sentences}, and in contrast to the F1 score, CoNES correctly returns a value of 1 as the entities are all present and all in the correct amounts.

\subsubsection{Use of socio-cultural information}

An important question when performing OCR correction using LMs is whether additional information about the text should be provided, such as cultural, social or historical information. Such information has typically not been the focus during the OCR or post-correction process. This section is to demonstrate the importance of including socio-cultural context in the prompt and how the task itself can stimulate the LM to create its own implicit prompt. To test this two controlled sub-experiments are performed, the first experiment checks the interaction between prompt and task, the second checks the effect of task length 

\subsubsection*{Prompt task interaction}

An OCR correction task is manually created to check the interaction between prompt and task by using three different, but related, prompts and tasks. The `corrupted' phrases are from the well-known joke `Why did the chicken cross the road? To get to the other side". As is shown in Table \ref{tab:jokes}, the joke is presented to the LM heavily corrupted using `*' with either only the setup of the joke, the punchline of the joke, or the full joke,
however, as is shown in Table \ref{tab:jokes}, they have been deliberately constructed such that this is not obvious. The the LM is given and provided one of three instructions prompts shown in Table \ref{tab:joke_prompts}, the prompts provide either basic instruction, basic instruction and additional socio-cultural context that the corrupted phrase is a joke, or basic prompt and misleading context that the text is from a cookery article. 
A misleading prompt is also a form of socio-cultural context but shapes the sampling distribution away from the socio-cultural-context, as such the misleading context should have worse performance than the baseline.
The three prompts and three phrases will produce nine scenarios. Although the task with both sentences is technically more complex due to the larger number of tokens that need to be returned correctly, the additional context provided by the task itself could make it easier for the LM to solve if the LM can draw on the broader socio-cultural context of jokes implicit in the text.

\begin{table*}[ht]
    \caption{True and corrupted phrases used to test the importance of socio-cultural context in OCR correction tasks.}
    \centering
    \begin{tabular}{p{0.17\linewidth}p{0.35\linewidth}p{0.35\linewidth}}
        \toprule
        Type & Corrupted Phrase &      Answer \\ \midrule
        setup & \verb|*** did the *** ***| \verb|*** ***|  & Why did the chicken cross the road?  \\ 
        punchline & \verb|*** *** *** ***| \verb|other side|  & To get to the other side  \\ 
        full  & \verb|*** did the *** ***| \verb|*** ***|  \verb|*** *** *** ***| \verb|other side|& Why did the chicken cross the road? To get to the other side  \\ 

        \bottomrule
    \end{tabular}
    \label{tab:jokes}
\end{table*}

\begin{table*}[ht]
    \caption{The three prompts used to test the importance of socio-cultural context}
    \centering
    \begin{tabular}{p{0.17\linewidth}p{0.7\linewidth}}
        \toprule
        Type & Prompt  \\ \midrule
        basic & Please correct the below sentences containing OCR errors    \\ 
        socio & Please correct the below sentences containing OCR errors, the sentences are part of popular jokes \\ 
        mislead  & Please correct the below sentences containing OCR errors, the sentences are part of an article on cookery \\ 

        \bottomrule
    \end{tabular}
    \label{tab:joke_prompts}
\end{table*}

Each of the nine prompt-joke combinations will be tested 100 times and evaluated per group, with a temperature of 0.8 to ensure some variation. Given the highly controlled nature of this experiment, the repetitions, and the ability to disentangle the effects of prompt and task, the evaluation metrics of the experiment metrics will be the fraction correct and Perplexity. Perplexity is a measure of uncertainty where higher values mean more uncertainty, as a result we will be able to interpret the impact of prompt on LM response in a more nuanced way than simply whether the response was correct or not. Perplexity is calculated using $\textrm{PP}(x) = e^{\overline{\textrm{ln}P(x)}}$ where $\overline{\textrm{ln}P(x)}$ is the mean of the log probability of the tokens of the LM response. The goal is to demonstrate in a controlled test that the socio-cultural context significantly impacts the outcome of the correction task. In addition, it will be clear if there is any interaction between the implicit information of the extended task and the prompt. As this is simply a conceptual demonstration, only GPT-4 will be used out of the seven LMs.

\subsubsection*{Task Length}

Having demonstrated the importance of the socio-cultural context, a more practical experiment will be performed. This experiment will take advantage of the line-aligned Sydney Morning Herald dataset. The experiment builds on the demonstration and tests two things: the whether increasing text length reduces overall error and the impact of socio-cultural context. In this experiment, three prompts will be used. The first prompt will be the `expert recover instructions' prompt described in table \ref{tab:combined_prompts}; the second prompt adds socio-cultural context with ``The text is from The Sydney Morning Herald 1842 -1950."; The third prompt provides a misleading socio-cultural context ``The text is from The Hong Kong Restaurant Review 1989-1993.". 
All 43 articles with at least 60 lines of text will be selected, but all text beyond line 60 will be removed to ensure a standard length and easy factorisation. The text will be broken up into blocks of $x$ rows, where x is one of the factors of 60 in the set $2, 5, 10, 15, 30, 60$, in other words, when $x=5$ each text is broken into 12 groups, whilst when $x = 30$ the text is broken into two groups. If the socio-cultural context is important, the longer the texts will perform better; in addition, the prompt which provides the socio-cultural context will perform the best, whilst the misleading prompt will perform the worst. Similar to the LM comparison, the performance of the different scenarios will be evaluated using CER.

\FloatBarrier
\section{Results}

The section is broken into three parts: Comparing the language models, The impact on the downstream tasks, and finally, the role of the socio-cultural context in the success of the LM at post-OCR correction.

\subsection{Comparison of Language Models}

In the comparison of language models, the results of the testing of the prompts on the development set were inconclusive  (see Supplementary Material section \ref{supp:prompt_testing}). A bootstrapped t-test showed there was no statistical difference between using the system message or putting the prompt after the text (p = 0.53). In addition the choice of prompt was not particularly clear cut, but prompts `expert recover publication text' and `full context' generally performed best across the models, as such these two prompts were chosen and placed after the text to be corrected.
The results of comparing the LMs on the two most high-performing prompts show that LMs are able to perform post-OCR correction. GPT-4 and Opus, the top performing LMs, reduced the CER by an error reduction percentage of over 60\% on the NCSE dataset (CER from 0.18 to 0.10) and obtained over 51\% on SMH (CER from 0.08 to 0.04), and 48\% on the CA dataset (CER from 0.1 to 0.05). Only Gemma and Mixtral could not improve on the baseline on any of the datasets; in contrast, Opus, GPT-4 and GPT-3.5 outperformed the baseline on all datasets. Table \ref{tab:results} shows performance across all models and datasets for CER and ERP for the full prompt. There were, however, significant differences in performance between the prompts on some LMs between the full and instruct prompts, but no overall pattern; see Supplementary Material \ref{supp:instruct_results} for the performance of the instruct prompt for each model. The performance of GPT-4 and GPT-3.5 was so strong that the prompts of \textcite{boros_post-correction_2024} were re-evaluated on all three datasets. This re-evaluation showed that the Boros prompt worked very well and was, in fact, one of the best-performing prompts (see Table \ref{tab:results}).

\begin{table*}
\caption{Model performance across the datasets measured in CER (Lower better) and Error Reduction Percentage (higher better).There is significant variation in how well the LMs are able to perform post-OCR correction, and significant differences between prompts for certain models.}
\label{tab:results}
\begin{tabular}{clrrrrrr}
\toprule
Model & \multicolumn{2}{c}{NCSE} & \multicolumn{2}{c}{SMH} & \multicolumn{2}{c}{CA} \\
 & CER & ERP & CER & ERP & CER & ERP \\
\midrule
Original & 0.17 & 0.00 & 0.08 & 0.00 & 0.10 & 0.00 \\
Claude 3 Haiku & 0.44 & 14.04 & 0.04 & 38.38 & 0.07 & 26.01 \\
Claude 3 Opus & 0.07 & 64.09 & 0.04 & 51.03 & 0.05 & 48.23 \\
GPT-3.5 & 0.23 & 37.65 & 0.04 & 39.18 & 0.06 & 44.22 \\
GPT-4 & 0.09 & 60.42 & 0.05 & 42.08 & 0.06 & 38.18 \\
GPT-4 Boros & 0.09 & 61.67 & 0.04 & 48.44 & 0.05 & 45.55 \\
Gemma 7B & 0.52 & -2.34 & 0.17 & -35.65 & 0.15 & -38.01 \\
Llama 3 & 0.40 & 19.12 & 0.07 & 12.90 & 0.11 & -9.42 \\
Mixtral 8x7B & 0.36 & 6.60 & 0.10 & -14.63 & 0.12 & -16.30 \\
Overproof & NaN & NaN & 0.05 & 28.38 & 0.07 & 34.59 \\
\bottomrule
\end{tabular}
\end{table*}

\begin{figure}
    \centering
    \includegraphics[width=\linewidth]{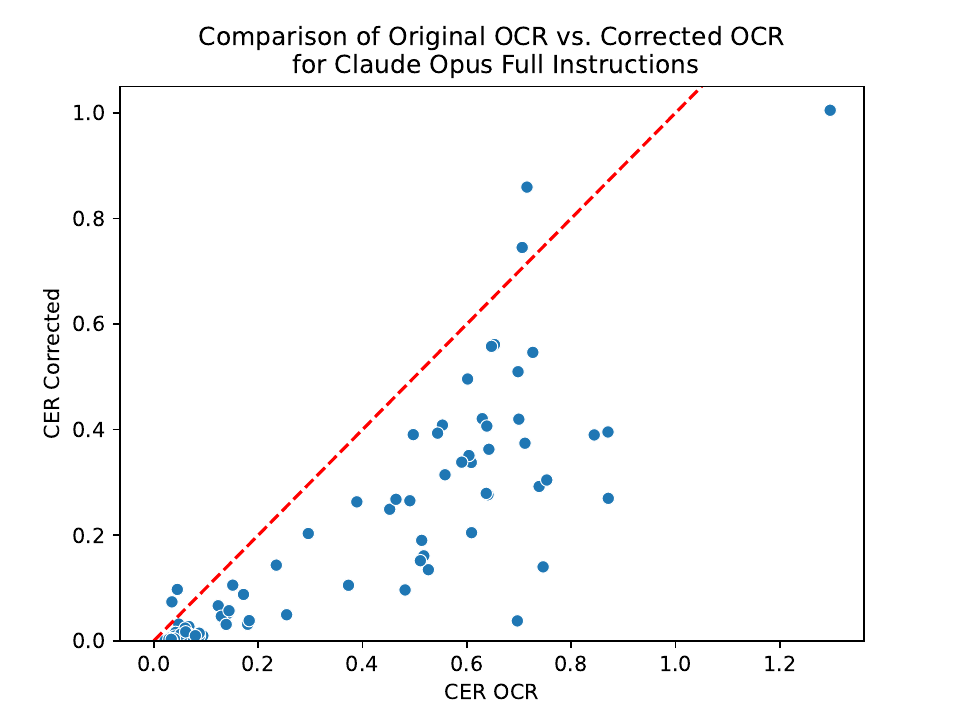}
    \caption{Relationship between the original and corrected CER using the Opus model. As the original CER gets increases so does the average corrected value. All texts below the red line have been improved by the CLOCR-C process.}
    \label{fig:cer_relationship}
\end{figure}

Figure \ref{fig:cer_relationship} shows the relationship between pre and post-correction CER using the top performing mode, Claude Opus. The figure shows that as the original CER increases the distribution and mean value of the corrected CER increases. Generally it seems that short list like articles such as recent book releases or theatre listings are recovered more easily than longer prose.

\subsection{NER Analysis}

The baseline CoNES scores for the raw OCR were not particularly high, with NCSE having 0.68, SMH having 0.86, and CA having 0.72. Analysing the impact of the correction on the downstream performance of Named Entity Recognition (NER) showed that all LMs except Gemma improved performance on all datasets relative to the baseline raw OCR; this was even in the case the LM made the error rate worse than the raw OCR. Opus performed exceptionally well, getting over 90\% in all three datasets. For completeness, the F1 scores were also calculated for the LMs (see supplementary material \ref{supp:NER}); however, for the reasons discussed in Section \ref{sect:meth_ner}, they are much less impressive, with generally only marginal performance increases.

\begin{table*}
\caption{CoNES performance across the three datasets shows that the best models create a significant improvement in data quality. However, the F1 erroneously shows poor recovery}
\label{tabl:ner_cones}
\begin{tabular}{clrrrrrr}
\toprule
Model & \multicolumn{2}{c}{NCSE} & \multicolumn{2}{c}{SMH} & \multicolumn{2}{c}{CA} \\
 & CoNES & F1 & CoNES & F1 & CoNES & F1 \\
\midrule
Raw OCR & 0.68 & 0.08 & 0.86 & 0.38 & 0.72 & 0.30 \\
Claude 3 Haiku & 0.72 & 0.11 & 0.95 & 0.46 & 0.85 & 0.35 \\
Claude 3 Opus & 0.92 & 0.22 & 0.97 & 0.53 & 0.92 & 0.40 \\
Gemma 7B & 0.59 & 0.09 & 0.84 & 0.24 & 0.62 & 0.18 \\
GPT-3.5 & 0.82 & 0.13 & 0.94 & 0.42 & 0.88 & 0.34 \\
GPT-4 Boros & 0.88 & 0.28 & 0.95 & 0.54 & 0.90 & 0.42 \\
GPT-4 & 0.88 & 0.19 & 0.95 & 0.32 & 0.89 & 0.36 \\
Llama 3 & 0.76 & 0.16 & 0.90 & 0.34 & 0.81 & 0.19 \\
Mixtral 8x7B & 0.68 & 0.08 & 0.91 & 0.21 & 0.78 & 0.13 \\
\bottomrule
\end{tabular}
\end{table*}

\subsection{Leveraging socio-cultural information}
\begin{figure}[htbp]
\centering
\includegraphics[width=0.45\textwidth]{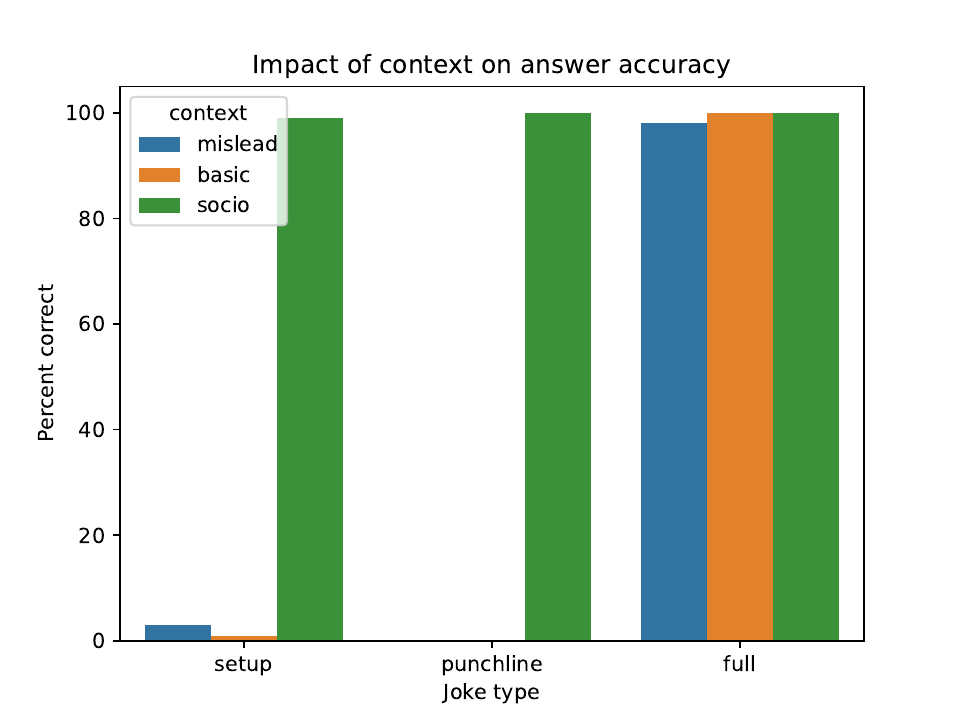}
\hfill 
\includegraphics[width=0.45\textwidth]{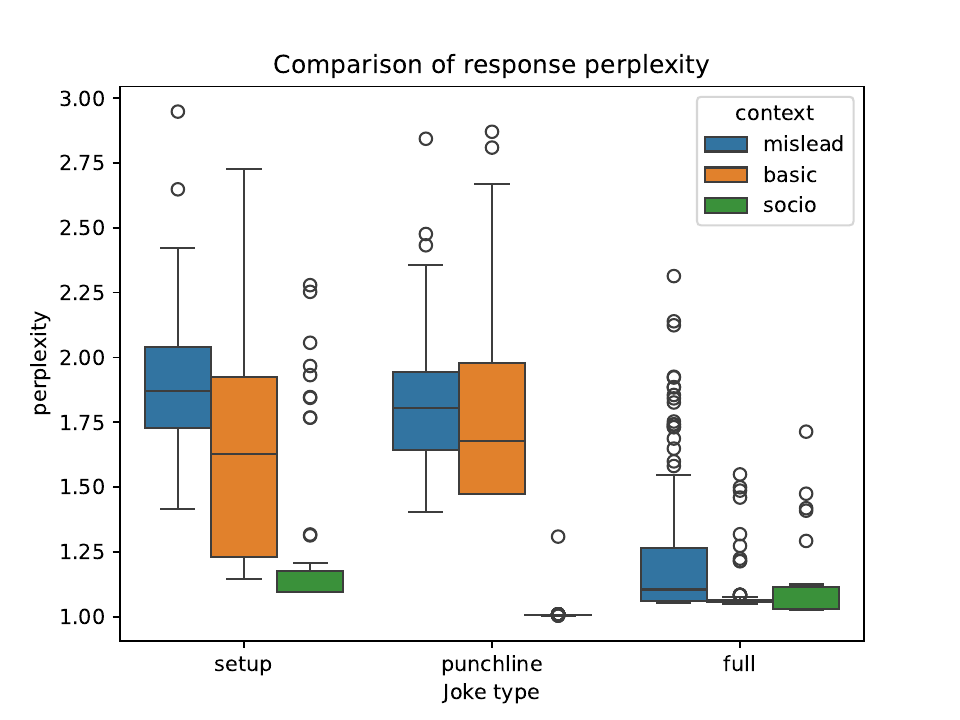}
\caption{The figures show that providing socio-cultural context in the prompt dramatically increases task performance.}
\label{fig:context_lev}
\end{figure}

Given that it is clear that LMs can perform post-OCR correction. It is interesting to know whether the socio-cultural context of the text can help with correction. The basic concept is demonstrated using the test case.
Figure \ref{fig:context_lev} demonstrates the impact of stimulating the LM (GPT-4) to use socio-cultural context. 
The setup and punchlines were almost never guessed when provided with the basic and misleading prompt, however were always guessed with the true socio-cultural context prompt. However, when the full joke was provided, even the misleading prompt correctly guessed the answer almost every time. It should be noted that this answer was often caveated with a comment along the lines of ``This may be a lighthearted introduction" or similar. The impact of the context can be seen on the perplexity, with Figure \ref{fig:context_lev} showing the median perplexity decreases as the context becomes more informative (or less misleading). For the combined prompt the perplexity of the misleading prompt is almost as low as the basic and true prompts. 

The experiment testing the interaction of task length and provided context is shown in Table \ref{tab:perf_lines}, the pattern is that the CER of the basic and socio-cultural prompts reduces as the task length increases. However, providing a misleading prompt creates substantially worse character error rate than the basic and socio-cultural prompts. This error is driven by GPT-4 observing the substantial mis-match between the prompt context and the actual text and returning a message highlighting this, the probability of producing this `error message' increases with the context length. The socio-cultural context marginally outperforms the basic prompt in all but task length 2 and 10, however, the difference is so small that it disappears when rounded to two decimal places.

\begin{table}
\caption{There is a clear relationship between the length of the task and overall performance. In addition it is clear that socio-cultural context in the prompt does have value, with misleading information performing worse in all tests, and the socio-cultural marginally outperforming the basic.}
\label{tab:perf_lines}
\begin{tabular}{rrrr}
\toprule
lines & basic & socio & mislead \\
\midrule
2 & 0.09 & 0.11 & 0.23 \\
5 & 0.03 & 0.03 & 0.17 \\
10 & 0.02 & 0.02 & 0.26 \\
15 & 0.03 & 0.02 & 0.37 \\
30 & 0.02 & 0.01 & 0.75 \\
60 & 0.01 & 0.01 & 0.83 \\
\bottomrule
\end{tabular}
\end{table}

\FloatBarrier
\section{Discussion}

The results of the experiments clearly show that LMs can be used for post-OCR correction. However, LM performance appears sensitive to the prompt, especially for short texts. Furthermore, it is unclear why some LMs are so effective at post-OCR correction whilst others are not, although the parameter size appears to be a factor as the largest models had the best performance. Reviewing the performance of the top model Claude Opus across the test showed that as expected recovery quality reduced as the CER of the OCR document increased, the model appears to perform better on list like articles as opposed to prose which is more prone to hallucination. Hallucination is particularly acute when there is substantial line mixing. In addition prose like articles seem to suffer more from LM commentary along the lines of `The article describes a tragic event'. It should be noted that the model appears perform some quite stunning reductions in error, these appear to be exclusively list like articles; efforts have been made to see whether the models have been trained on the text using another dataset but no evidence that this is the case has been found.

The poor performance shown in \textcite{boros_post-correction_2024}, was found not to be due to their prompt. Whilst part of the reason for this poor performance may be due to newer versions of GPT-3.5 and GPT-4, an investigation of the code used in \cite{boros_post-correction_2024} suggests that they used the mean instead of the median. To test the impact of the use of the mean as a metric the results of this paper were re-averaged using the mean instead of the median. It showed that CER got worse for almost all models, despite the majority of documents being improved by the CLOCR-C process (See supplementary material \ref{supp:mean_results}). These findings show the importance of careful metric choice and awareness of the type of distributions found in the data.

Analysing the impact of the post-OCR correction on the downstream tasks showed that LMs that did well on the post-OCR correction task also performed well on improving the CoNES score, with several models getting over 90\% similarity with the ground truth on at least one of the datasets. This improvement shows that post-OCR correction can mitigate some of the concerns of \cite{chiron_impact_2017, van_strien_assessing_2020} when it comes to the ability to identify entities or successfully search text databases created from OCR documents. The difference in the apparent model performance when comparing CoNES and F1 shows the weakness of using F1 for NER when considering the effectiveness of post-OCR correction techniques due to the likely different absolute text positions of any entities.

A simple controlled demonstration showed that providing a socio-cultural context to a prompt can dramatically improve task performance on post-OCR correction.
When the setup and punchline were combined into a single task, even the misleading prompt got 98\% accuracy. Moreover, the mean perplexity of the combined responses suggests that the implicit context from the task itself was sufficient, and the additional context provided no further value. The importance of the socio-cultural context was supported in the follow-up experiment, which showed that the error decreased as the length of the supplied text increased. 
It was interesting to note that the whole joke was recovered with the misleading prompt, but newspaper text was much less likely to be recovered with the misleading prompt, particularly as text length increased. Reading the messages from the LM suggests that it may be that once the divergence between the prompted context and the context provided by the task increases, the LM first produces a ``warning" and then an ``error message".
The results on socio-cultural context suggest that understanding a text's socio-cultural background can enhance correction quality. However, they also reveal that the practical value of detailed socio-cultural context in real tasks is somewhat ambiguous. This is because detailed prompts may be less beneficial than task length, as the language model (LM) can infer context from the task itself. Such a process could be described as a variant of In Context Learning or Many Shot In Context Learning \cite{agarwal_many-shot_2024} as the task implicitly supplies the context required to solve itself. This Task, Inferred In Context Learning (TIICL), is important to consider in relation to post-OCR correction as creating very specific prompts may be less important for longer or less corrupted text as the necessary contextual information is available from the text itself.

CLOCR-C is unlikely to be an explicitly trained behaviour of the LMs used in this paper. Therefore, it is reasonable to describe it as an emergent behaviour. However, it is not easy to explain why the behaviour has emerged; number of parameters alone is not sufficient because, as was shown by \textcite{boros_post-correction_2024}, previous versions of GPT-3.5 and GPT-4 did not beat the baseline. Perhaps there has been some development in the training regime, such as the recently developed multi-token prediction \cite{gloeckle_better_2024}, or a subtle shift in the data. However, without some new insight into how the models are trained, it is not possible to know.

One of the primary limits of the study is that post-OCR correction has to work within the confines of the actual OCR process. As was shown in Table \ref{tab:example_text}, the scanning process can make errors in the physical layout of the page, such as mixing printed lines, poor segmentation of articles and failing to separate columns. Such errors resulted in the median being used to prevent a massive error on a single article skewing the distribution of results. Post-OCR correction will always struggle with errors of physical position and, as such, requires at least this aspect of the OCR process to be relatively good. 

\section{Conclusion}

This paper set out to test whether LMs can be used to perform post-OCR correction. The paper also showed that providing the socio-cultural context of the text on which post-OCR correction was performed improved performance while providing misleading prompts lowered performance. Given these findings, the term Context Leveraging OCR-Correction (CLOCR-C) seems appropriate as the LM leverages both the broader socio-cultural context in the prompt as well as what appears to be a form of Task inferred In Context Learning, which reduces the impact of the prompt as the text to be corrected gets longer.
However, it should be noted that in common with the findings of \textcite{boros_post-correction_2024}, not all LMs successfully reduced the error rate, with some LMs substantially increasing error. Although the top-performing models were very effective at CLOCR-C, the cost of using a large closed-source model to correct a digital archive is likely prohibitively expensive. This limitation highlights the need for further work focused on training open-source models that could be deployed much more cheaply, making CLOCR-C a more accessible solution for a broader range of applications. Another area that would benefit from further investigation would be a method to predict if CLOCR-C will be able to recover text or whether it is too corrupted. Overall this paper shows CLOCR-C is a promising approach to OCR correction that takes advantage of the ability of the LMs to leverage the socio-cultural information provided in the prompt and to perform Task Induced In Context Learning using the OCR text itself.

\section*{Data and Code}

The transcribed dataset is available at the UCL data repository \url{https://rdr.ucl.ac.uk/articles/dataset/Transcribed_newspaper_articles_from_the_NCSE_collection/25805008}. All code is available at \url{https://github.com/JonnoB/clocrc}

\section*{Acknowledgments}

I would like to thank the NCSE team and the King's Digital Laboratory for providing access to the data and support in interpreting it. I would also like to thank Anindya Basu and James Hale for their thoughtful and constructive feedback on the manuscript.

\printbibliography

\appendix

\section{Prompt testing}
\label{supp:prompt_testing}
\begin{figure}[ht]
\centering
\begin{tabular}{cc}

\begin{minipage}{0.45\textwidth}
\includegraphics[width=\linewidth]{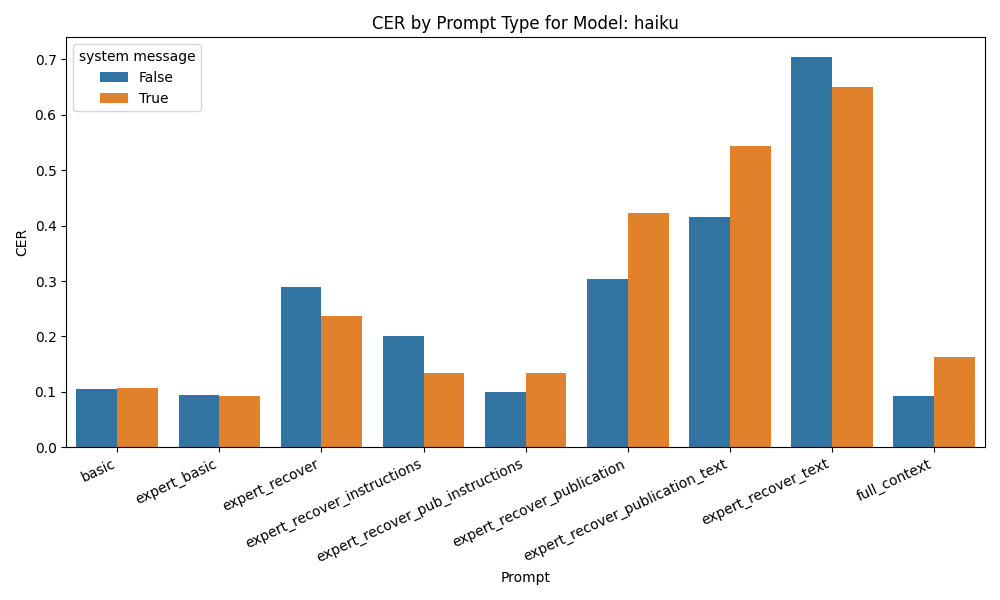} 
\end{minipage} &

\begin{minipage}{0.45\textwidth}
\includegraphics[width=\linewidth]{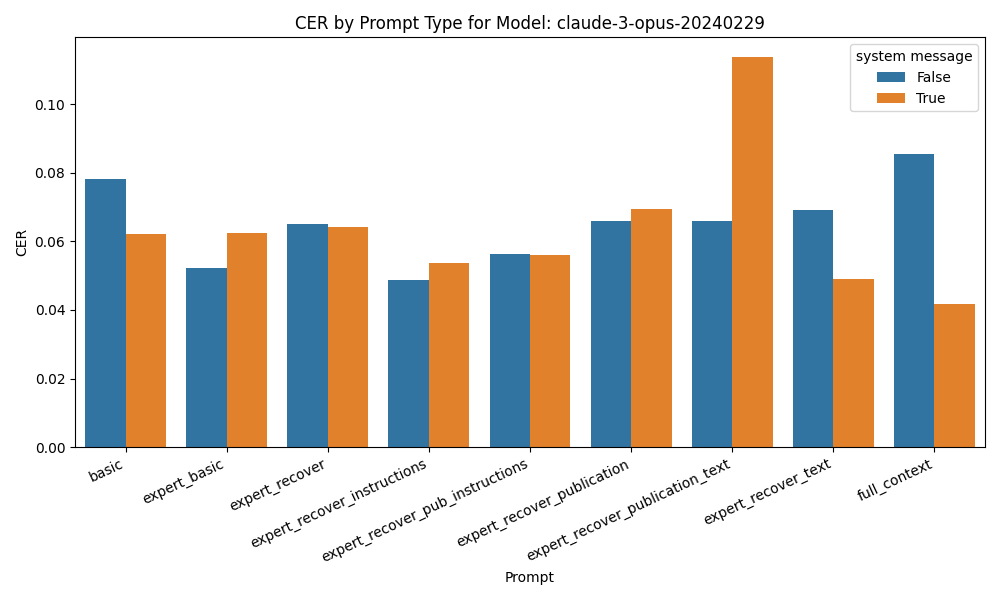}
\end{minipage} \\

\begin{minipage}{0.45\textwidth}
\includegraphics[width=\linewidth]{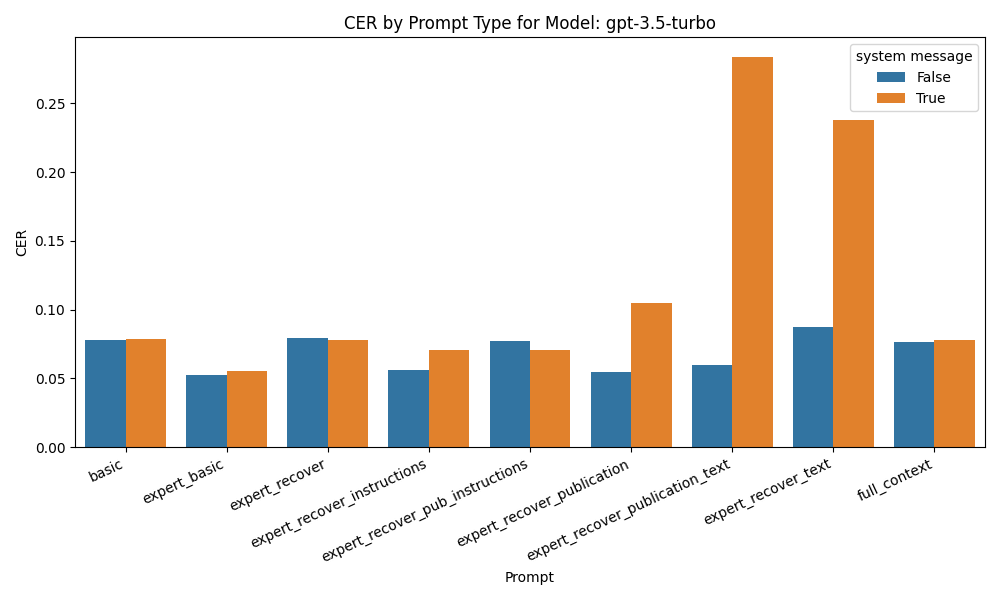}
\end{minipage} &

\begin{minipage}{0.45\textwidth}
\includegraphics[width=\linewidth]{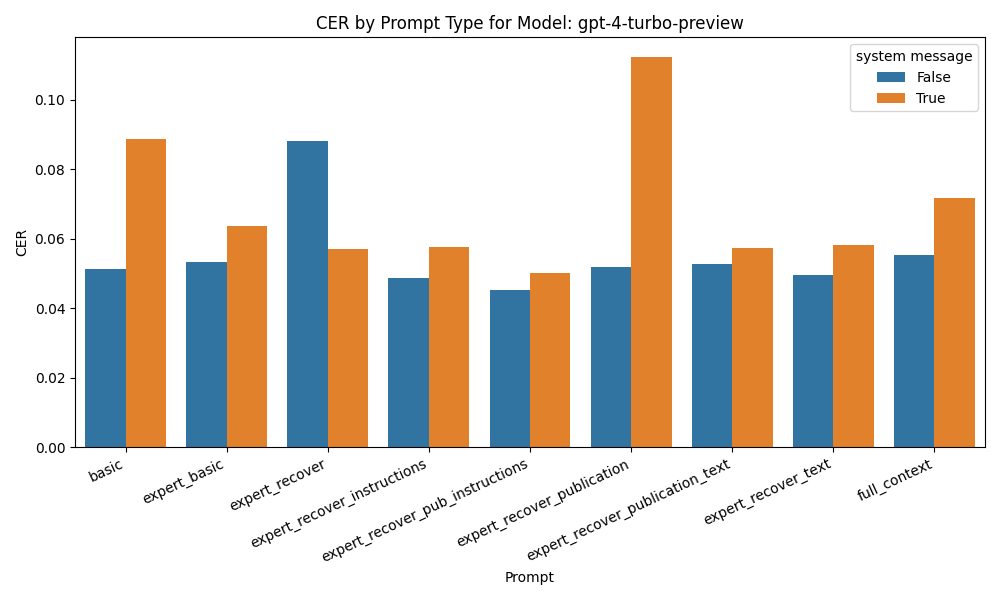}
\end{minipage}

\end{tabular}
\caption{The overall results are not clear with system prompts working sometimes and not others, and prompts working well for some models but not others} 
\label{fig:prompt_test}
\end{figure}

The prompt test, shown in Figure \ref{fig:prompt_test}, was not very conclusive. The test comparing using the system prompt or the prompt after the text was relatively evenly split between the two options, and a bootstrapped t-test did not return significance. In addition, there was little agreement between the prompts, with some prompts performing well on certain models but poorly on others. Overall, the full prompt generally did well, as did The prompts 'expert\_recover\_pub\_instructions' (b+c+d+e) and `expert\_recover\_instructions' (b+c+f). As the paper is a proof of concept and not an optimisation task, the decision was taken to use text followed by prompt, and for the choice of prompt, the full prompt as well as `expert\_recover\_instructions'.

\section{Instruct results}
\label{supp:instruct_results}
This section shows the results of the prompts using the `expert\_recover\_instructions' (b+c+f), and compares them with the full instructions across the models. As when using the full prompt, Opus dominates outperforming all other models.

\begin{table}
\caption{Model performance across the datasets measured in Error Reduction Percentage, higher is better.There is significant variation in how well the LMs are able to perform post-OCR correction, and significant differences between prompts for certain models.}
\label{tab:results}
\begin{tabular}{p{5cm}cccc}
\toprule
Model & NCSE & SMH & CA \\
\midrule
Claude 3 Haiku & 27.80 & 35.70 & 34.70 \\
Claude 3 Opus & \textbf{62.70} & \textbf{45.50} & \textbf{47.00} \\
GPT-3.5 & 39.40 & 42.90 & 44.10 \\
GPT-4 & 59.80 & 41.80 & 37.60 \\
Gemma 7B & 0.10 & -12.90 & -41.00 \\
Llama 2 70B & -11.20 & 6.50 & -6.50 \\
Llama 3 & 16.10 & 17.90 & 14.20 \\
Mixtral 8x7B & 7.00 & -19.10 & -22.10 \\
\bottomrule
\end{tabular}
\end{table}

\subsection{Comparison between the `Full' and `Instruct' prompts}

\begin{figure}
    \centering
    \includegraphics[width = \textwidth]{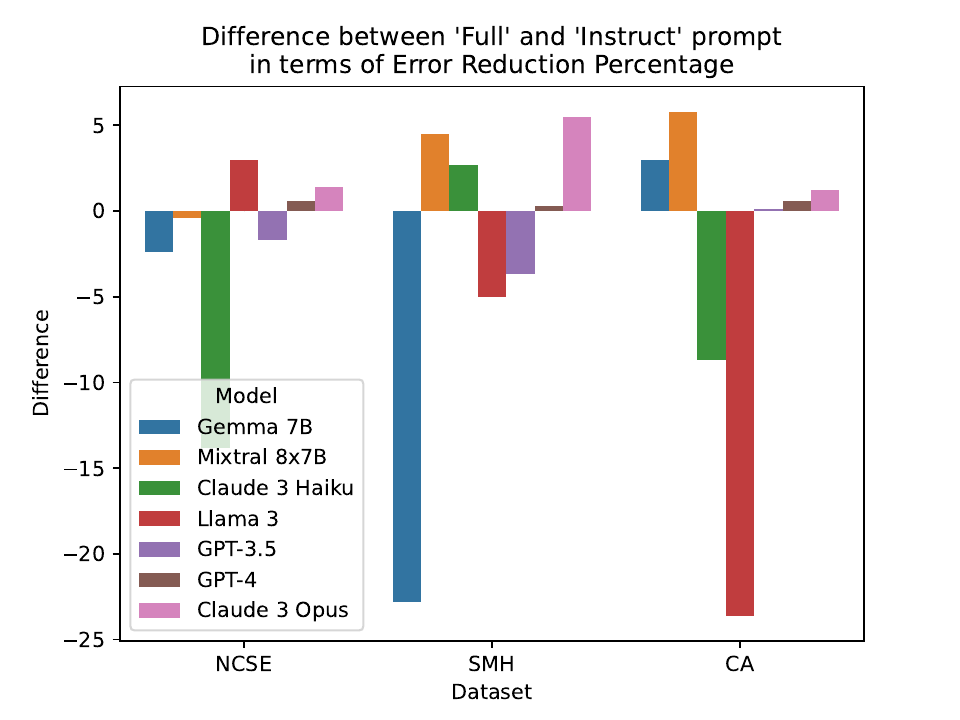}
    \caption{The difference between the instruct prompt and the full prompt is different across models and datasets showing no clear picture.}
    \label{fig:prompt_diff}
\end{figure}

Much like the original prompt testing, there was no clear result when comparing the results of the 'Full' and 'Instruct' prompts across the three datasets. As shown in Table \ref{tab:prompt_diff} and Figure \ref{fig:prompt_diff}, the results varied across both model and dataset.

\begin{table}
\caption{The difference between the two prompts across all models and datasets}
\label{tab:prompt_diff}
\begin{tabular}{p{5cm}cccc}
\toprule
Model & Dataset & Full & Instruct & Difference \\
\midrule
Claude 3 Haiku & NCSE & 14.00 & 27.80 & -13.80 \\
Claude 3 Haiku & CA & 26.00 & 34.70 & -8.70 \\
Claude 3 Haiku & SMH & 38.40 & 35.70 & 2.70 \\
Claude 3 Opus & CA & 48.20 & 47.00 & 1.20 \\
Claude 3 Opus & NCSE & 64.10 & 62.70 & 1.40 \\
Claude 3 Opus & SMH & 51.00 & 45.50 & 5.50 \\
GPT-3.5 & CA & 44.20 & 44.10 & 0.10 \\
GPT-3.5 & NCSE & 37.70 & 39.40 & -1.70 \\
GPT-3.5 & SMH & 39.20 & 42.90 & -3.70 \\
GPT-4 & NCSE & 60.40 & 59.80 & 0.60 \\
GPT-4 & CA & 38.20 & 37.60 & 0.60 \\
GPT-4 & SMH & 42.10 & 41.80 & 0.30 \\
Gemma 7B & CA & -38.00 & -41.00 & 3.00 \\
Gemma 7B & NCSE & -2.30 & 0.10 & -2.40 \\
Gemma 7B & SMH & -35.70 & -12.90 & -22.80 \\
Llama 3 & NCSE & 19.10 & 16.10 & 3.00 \\
Llama 3 & CA & -9.40 & 14.20 & -23.60 \\
Llama 3 & SMH & 12.90 & 17.90 & -5.00 \\
Mixtral 8x7B & CA & -16.30 & -22.10 & 5.80 \\
Mixtral 8x7B & NCSE & 6.60 & 7.00 & -0.40 \\
Mixtral 8x7B & SMH & -14.60 & -19.10 & 4.50 \\
\bottomrule
\end{tabular}
\end{table}

\section{NER Analysis: F1 score}
\label{supp:NER}
\begin{figure}
    \centering
    \includegraphics[width = \textwidth]{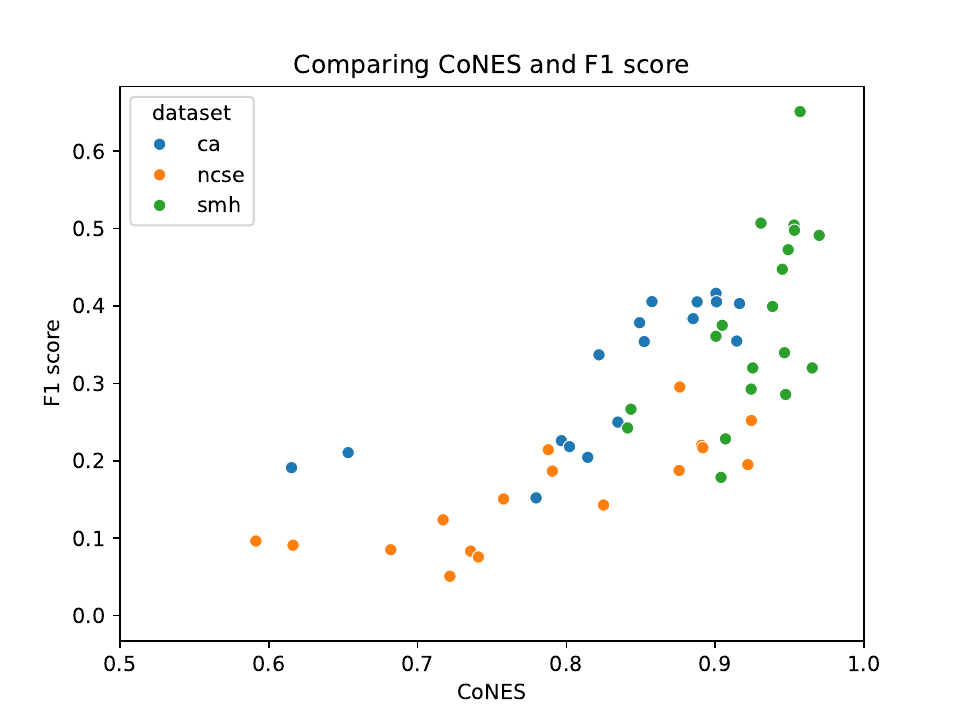}
    \caption{The relationship between CoNES and F1 improvement is substantial, with every model having a higher CoNES improvement than F1 improvement. }
    \label{fig:CoNES_vs_F1}
\end{figure}

The baseline F1 scores for the raw OCR were very low, with NCSE having 0.08, SMH having 0.43, and CA having 0.28. As can be seen in the paper, all models beat the respective dataset baseline. However, as shown in Figure \ref{fig:CoNES_vs_F1} the overall score was much lower than compared with the CoNES score. This difference is largely due to the issue of the corrected text having lengths different from the ground truth. As a result, the position of the Named Entities is outside the allowed window so even a correct match does not count as the position is wrong. This issue does not affect CoNES as it is position agnostic.

\section{Mean performance of the models}
\label{supp:mean_results}

Table \ref{tab:results_mean} shows the performance of the models when using the mean instead of the median. In the majority of cases the performance is worse, sometimes substantially. This is because when the LM's make errors they can make really bad errors, including completely hallucinated results and repeating words or phrases over and over again. However, generally they perform very well. This sort of behaviour naturally leads to skewed results making the mean an inappropriate metric, and likely explaining the poor results found in \cite{boros_post-correction_2024}.

\begin{table}
\caption{When the mean is used instead of the median almost all results are worse, in some cases substantially so, despite the majority of articles being improved.}
\label{tab:results_mean}
\begin{tabular}{clrrrrrr}
\toprule
Model & \multicolumn{2}{r}{NCSE} & \multicolumn{2}{r}{SMH} & \multicolumn{2}{r}{CA} \\
 & CER & ERP & CER & ERP & CER & ERP \\
\midrule
Claude 3 Haiku & 0.40 & -237.41 & 0.11 & -40.65 & 0.18 & -92.14 \\
Claude 3 Opus & 0.27 & -28.51 & 0.05 & 45.92 & 0.06 & 43.51 \\
GPT-3.5 & 0.27 & -42.03 & 0.09 & -0.52 & 0.11 & -18.29 \\
GPT-4 & 0.17 & 58.03 & 0.06 & 32.65 & 0.07 & 29.70 \\
GPT-4 Boros & 0.17 & 60.32 & 0.05 & 44.01 & 0.06 & 44.23 \\
Gemma 7B & 0.46 & -241.18 & 0.27 & -227.08 & 0.29 & -209.20 \\
Llama 3 & 0.36 & -180.80 & 0.18 & -111.94 & 0.20 & -108.12 \\
Mixtral 8x7B & 0.37 & -113.22 & 0.17 & -58.17 & 0.17 & -60.65 \\
Overproof & NaN & NaN & 0.07 & 27.28 & 0.07 & 34.11 \\
\bottomrule
\end{tabular}
\end{table}

\end{document}